\documentclass[conference]{IEEEtran}
\IEEEoverridecommandlockouts

\usepackage[hidelinks]{hyperref}
\usepackage{graphicx}%
\usepackage{multirow}%
\usepackage{amsmath,amssymb,amsfonts}%
\usepackage{amsthm}%
\usepackage{mathrsfs}%
\usepackage{xcolor}%
\usepackage{textcomp}%
\usepackage{booktabs}%
\usepackage{algorithm}%
\usepackage{algorithmicx}%
\usepackage{algpseudocode}%
\usepackage{listings}%
\usepackage{array}

\usepackage{color, colortbl}
\definecolor{LightCyan}{rgb}{0.88,1,1} 

\usepackage{subcaption}

\def\BibTeX{{\rm B\kern-.05em{\sc i\kern-.025em b}\kern-.08em
T\kern-.1667em\lower.7ex\hbox{E}\kern-.125emX}}
\begin{document}
\title{RegMix: Adversarial Mutual and Generalization Regularization for Enhancing DNN Robustness\thanks{We thank the support of the EPSRC-funded project National Edge AI Hub for Real Data: Edge Intelligence for Cyber-disturbances and Data Quality (EP/Y028813/1).}
}

\author{
\IEEEauthorblockN{Zhenyu Liu}
\IEEEauthorblockA{\textit{School of Computing} \\
\textit{Newcastle University}\\
Newcastle, United Kingdom \\
z.liu48@newcastle.ac.uk}
\and
\IEEEauthorblockN{Varun Ojha}
\IEEEauthorblockA{\textit{School of Computing} \\
\textit{Newcastle University}\\
Newcastle, United Kingdom \\
varun.ojha@newcastle.ac.uk}
}
%
%
%
%

%
\maketitle
\begin{abstract}
Adversarial training is the most effective defense against adversarial attacks. The effectiveness of the adversarial attacks has been on the design of its loss function and regularization term. The most widely used loss function in adversarial training is cross-entropy and mean squared error (MSE) as its regularization objective. However, MSE enforces overly uniform optimization between two output distributions during training, which limits its robustness in adversarial training scenarios. To address this issue, we revisit the idea of mutual learning (originally designed for knowledge distillation) and propose two novel regularization strategies tailored for adversarial training: (i) \textit{weighted adversarial mutual regularization} and (ii) \textit{adversarial generalization regularization}. In the former, we formulate a decomposed adversarial mutual Kullback–Leibler divergence (KL-divergence) loss, which allows flexible control over the optimization process by assigning unequal weights to the main and auxiliary objectives. In the latter, we introduce an additional clean target distribution into the adversarial training objective, improving generalization and enhancing model robustness. Extensive experiments demonstrate that our proposed methods significantly improve adversarial robustness compared to existing regularization-based approaches. Our code is available https://github.com/lusti-Yu/Regmix.git.

\end{abstract}
%
\section{Introduction}\label{sec:intro}
Loss functions are fundamental in training deep neural networks that guide model optimization and generalization. Among them, cross-entropy (CE) loss is a widely adopted loss function in standard image classification tasks, and CE is a core objective in adversarial training frameworks too~\cite{shafahi2019adversarial}. In addition, mean squared error (MSE), although traditionally used in regression tasks, has demonstrated effectiveness in improving robustness under adversarial training settings~\cite{cui2021learnable}. Another widely used loss function is the Kullback-Leibler divergence (KL-divergence), which measures the divergence between a predicted and target distribution. KL-divergence serves as a key loss function in many applications, including knowledge distillation~\cite{zhang2018deep} and adversarial knowledge distillation~\cite{huang2023boosting}, due to its ability to capture fine-grained differences between probability distributions. To bridge this gap, theoretical analyses~\cite{cui2024decoupled} have shown that the KL-divergence loss can be decomposed into a weighted combination of the MSE and CE.

A recent approach, FGSM-PGK~\cite{jia2024improving}, introduced an effective regularization method to enhance model robustness by minimizing the squared distance (\(\ell_2 \)-norm) between the adversarial prediction obtained through prior-guided initialization (which incorporates information from adversarial examples generated in previous epochs) and the final adversarial prediction. Notably, the \(\ell_2 \)-norm regularization term requires backpropagation through both the initial and final adversarial outputs. Similarly, mutual learning~\cite{zhang2018deep} has been proposed in previous knowledge distillation works, utilizing Kullback-Leibler (KL) divergence for bidirectional alignment between the output distributions of two models. In the context of \textit{mutual learning}, the decomposed formulation of KL divergence allows intuitive control over the trade-off between the two terms by adjusting their respective weights. However, achieving a similar trade-off operation in single-model settings with a single \(\ell_2 \)-norm term is more challenging. To address this limitation, we draw inspiration from the asymmetry of the KL-divergence, as explored in mutual learning~\cite{zhang2018deep} within the context of adversarial training. Building upon this idea, we incorporate a symmetric KL-divergence-based regularization into adversarial training for single-model settings. This design aims to improve model robustness by embedding a pair of symmetric KL-divergence terms into the adversarial loss function. Furthermore, we observe that generalization-aware adversarial training on adversarial outputs contributes positively to robustness enhancement.
\begin{figure}[t]
    \centering
    \includegraphics[width=0.98\linewidth]{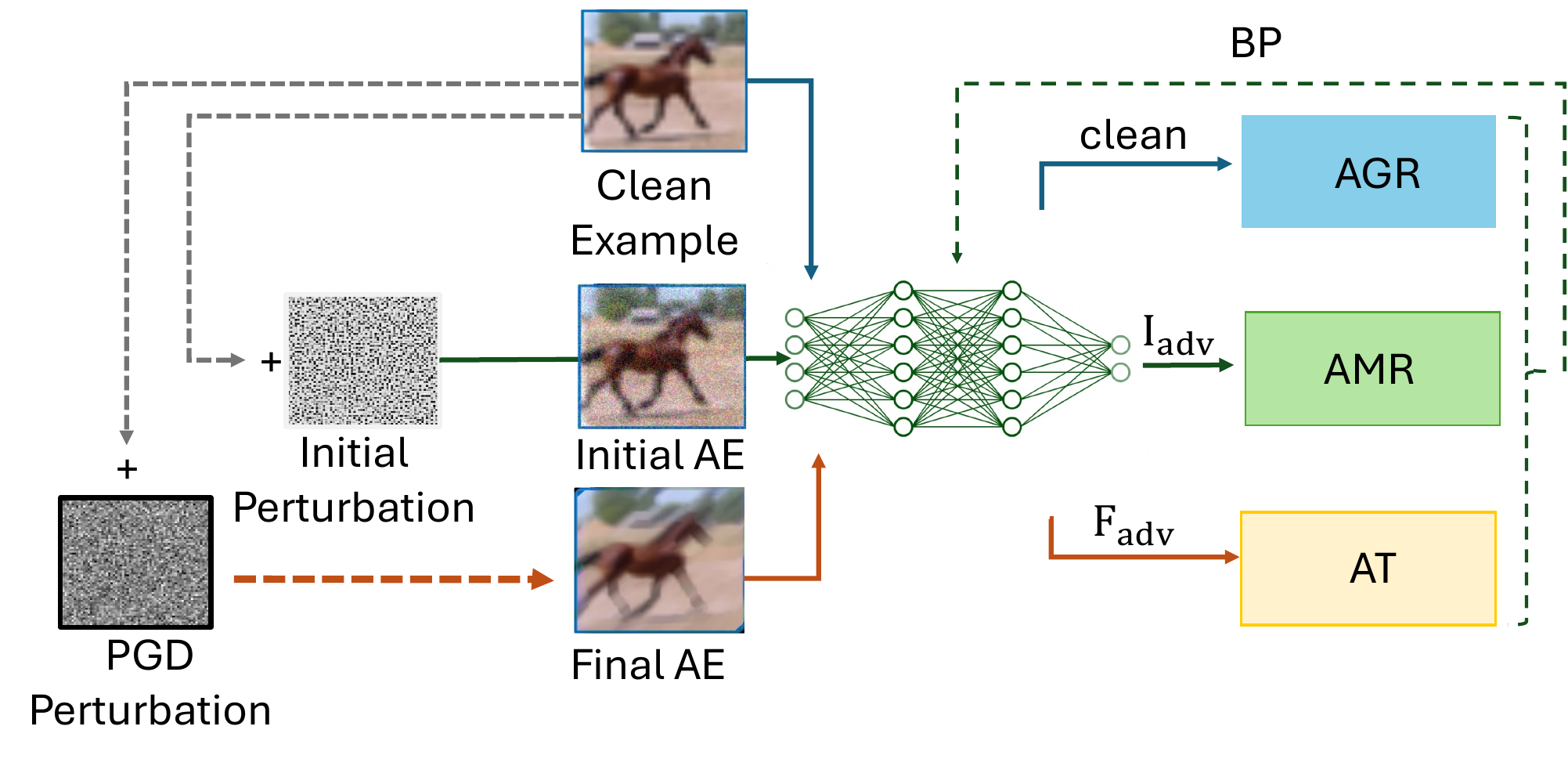} 
    \caption{
        An overview of the proposed RegMix framework. 
        AE denotes the adversarial example, BP stands for backpropagation, and CLN represents the clean output. 
        \(I_{\text{adv}}\) and \(F_{\text{adv}}\) indicate the initial (randomly initialized perturbation without gradient) and final (gradient-based) adversarial outputs, respectively. 
        \(y\) is the ground-truth label. 
        The two core components of our method, Adversarial Mutual Regularization (AMR) and Adversarial Generalization Regularization (AGR), are introduced in Sections~\ref{Adversarial mutual} and~\ref{Adversarial Generalization}, respectively. 
        AT refers to adversarial training.
    }
    \label{fig:regmix}
\end{figure}


Based on the above observations~\cite{zhang2018deep}\cite{jia2024improving}, we propose RegMix (adversarial mutual and generalization regularization) for enhancing deep neural networks (DNN) robustness (see Fig.~\ref{fig:regmix}). RegMix consists of two regularization strategies for the single-model setting: (i) \textit{mutual adversarial regularization} and (ii) \textit{adversarial generalization regularization}. In particular, we replace the conventional $\ell_2$-norm with a mutual adversarial learning objective, formulated as an asymmetric loss function with two unequally weighted components. The proposed mutual adversarial learning regularization consists of a pair of symmetric KL divergence terms applied between the adversarial and prior-initialized adversarial predictions. Due to the asymmetry of the KL divergence, the two directions of approximation convey different meanings. Therefore, we prioritize the approximation from the final adversarial output to the initial adversarial output as the main objective, while treating the reverse direction as auxiliary. In addition, we introduce a generalization-based training approach on adversarial outputs to further improve model robustness. Specifically, we integrate the clean output distribution into the regularization term of adversarial training and propose a mixed-distribution adversarial training strategy. This regularization aims to increase the diversity of the target distribution, thereby enabling adversarial training to enhance generalization and ultimately leading to improved robustness.

Our experimental findings suggest that employing our proposed KL divergence loss function in adversarial training not only enhances robustness against standard perturbations but also yields notable improvements when facing stronger adversarial attacks. Our main contributions are in the following aspects:




\begin{itemize}

    \item We introduce Adversarial Mutual Regularization (AMR), which employs a pair of symmetric weighted KL (wKL) losses, comprising a primary and an auxiliary term with different weights.
    
    \item We introduce a training strategy, Adversarial Generalization Regularization (AGR), which incorporates generalized target distributions into adversarial training to enhance model robustness.

    \item We discover that training with probability distributions not only enhances robustness against adversarial attacks of similar scale during training but also improves defense against stronger attacks.
\end{itemize}

The organization of the rest of the paper is as follows. Sec.~\ref{sec:related_work} discusses works relevant to adversarial training and $\ell$-norm loss in adversarial training. Our RegMix methodology is discussed in Sec.~\ref{sec:method} followed by experiments and results in Sec.~\ref{sec:exp_results}. We conclude in Sec.~\ref{sec:con} that the RegMix methodology not only enhances robustness against adversarial attacks of similar scale during training but also improves defense against stronger attacks. 

\section{Related work}
\label{sec:related_work}
\subsection{Adversarial attacks for robustness evaluation.}  
Among the existing adversarial attack methods, the following are most commonly used to evaluate the robustness of target models. FGSM~\cite{goodfellow2014explaining}, as one of the most classic adversarial attack techniques, generates adversarial examples in a single step based on the model's gradient under the white-box setting. Following this, the more powerful PGD~\cite{madry2017towards} attack was proposed, which also utilizes gradients but performs multiple iterative updates to generate stronger adversarial perturbations. Carlini and Wagner (C\&W)~\cite{carlini2017towards} further introduced the C\&W attack, a highly effective optimization-based method. In addition, the AutoAttack (AA)~\cite{andriushchenko2020square,croce2020minimally} framework combines several strong attacks and is widely used as a comprehensive and reliable robustness evaluation benchmark. These attacks are generally conducted under white-box settings to comprehensively assess the model's performance against various types of adversarial perturbations.

\subsection{Adversarial training} 
Adversarial training is an effective approach to enhancing model robustness, with multi-step Projected Gradient Descent (PGD) attacks being widely adopted in adversarial training~\cite{rice2020overfitting} and demonstrating significant robustness improvements. However, the computational cost of multi-step PGD for adversarial training remains high. To address this issue, single-step adversarial training methods have been proposed \cite{andriushchenko2020understanding}, \cite{kim2021understanding}, \cite{sriramanan2021towards}, where the Fast Gradient Sign Method (FGSM) is commonly used. Although FGSM-based adversarial training (FGSM-AT~\cite{wong2020fast}) is computationally efficient and improves model robustness, it suffers from catastrophic overfitting. To maintain low computational cost while mitigating catastrophic overfitting, PGD-2-AT was introduced, balancing efficiency and robustness to some extent.  Furthermore, FGSM-PGK~\cite{jia2024improving} incorporates PGD-2-AT~\cite{jia2022prior} and leverages regularization techniques to further enhance robustness. Additionally, the AD method emphasizes the importance of probability distributions and improves robustness using the KL-divergence. Each of these methods has its own advantages.  

\subsection{Applications of norm loss in adversarial training}
The MSE loss has recently gained attention as an effective component in adversarial training frameworks. In LBGAT~\cite{cui2021learnable}, the target model is trained solely with a single MSE loss to guide adversarial learning. Sriramanan et al. proposed GAT~\cite{sriramanan2020guided}, which introduces a nuclear-norm-based smoothing guided regularization to enhance robustness. FGSM-PGI~\cite{jia2022prior} incorporates MSE as a regularization term combined with sample initialization, and FGSM-PGK~\cite{jia2024improving} further improves robustness on this basis.

While MSE-based methods have shown promising results, some approaches use a single MSE loss to optimize two output targets simultaneously. This shared objective may limit task-specific learning and reduce the overall robustness of the model. Moreover, knowledge distillation from a teacher model to a student model commonly utilizes the KL-divergence to align output distributions. Inspired by these advancements and limitations, we propose a novel regularization strategy that introduces probability distribution constraints to enhance robustness.

\section{RegMix Methodology}
\label{sec:method}
This section is organized as follows. First, we revisit previous methods and describe how our observations lead to the consideration of whether modifying the initial distribution is necessary in Sec.~\ref{Preliminary}. We then investigate how to effectively assign and balance the decomposed primary and auxiliary KL terms, as detailed in Sec.~\ref{Adversarial mutual}, and propose a generalization target distribution method in Sec.~\ref{Adversarial Generalization}.

\subsection{Preliminary}
\label{Preliminary}
In adversarial training, both FGSM-PGI~\cite{jia2022prior} and FGSM-PGK~\cite{jia2024improving} leverage regularization losses, which play a crucial role in enhancing robustness. Specifically, each mini-batch contains both initial adversarial outputs, which are sampled from the clean output, and final adversarial outputs, which are obtained directly from the model's adversarial output. At each time step \( t \), the model receives a mini-batch of data \( \{ (x_i^{(t)}, y_i^{(t)}) \}_{i=1}^{n_t} \), where \( n_t \) denotes the number of samples in the batch. The function \( f(\mathbf{x} + \boldsymbol{\delta}^{\text{initial}}) \) represents the model's output under the initial adversarial perturbation, while \( f(\mathbf{x} + \boldsymbol{\delta}^{\text{final}}) \) corresponds to the output under the final adversarial perturbation. The goal of regularization is to update the model through the initial and final adversarial outputs.  Given a model \( f_{\boldsymbol{w}} \) with learnable parameters \( {\boldsymbol{w}} \),  symbol \( \mathcal{L} \) indicate model's regularization loss function.

PGD-2-AT is designed as a fast and effective adversarial training method, addressing the rapid robustness degradation observed in FGSM-RS and FGSM-AT. Specifically, it constructs a regularization term based on the prior-guided initial output and the adversarial output. The objective of this term in adversarial training is to enforce similarity between the two adversarial outputs. Methodologically, it employs the $\ell_2$ distance to constrain these outputs:  
\begin{equation}
    \mathcal{L} (x,y)=\left\|f_{\boldsymbol{w}}(\boldsymbol{x}+\boldsymbol{\delta}^{final}) -f_{\boldsymbol{w}}(\mathbf{x}+\boldsymbol{\delta}^{initial})\right\|^2.
\end{equation}  

In such cases, the two terms in the same loss formulation are treated as equally important during training. However, we contend that treating both components equally may limit the potential of the model to reach its optimal performance. Here, \(\boldsymbol{\delta}^{initial}\) is a perturbation randomly initialized without gradient involvement, while \(\boldsymbol{\delta}^{final}\) is generated by PGD with gradient guidance. These two perturbations represent different aspects of the model's robustness: one without a gradient and one with a gradient.

Our key assumptions are summarized as follows: Firstly, the two adversarial outputs may not contribute equally to the adversarial training process, making it possible to identify which one should play the primary role and which should act as a supportive component to maximize the effectiveness of adversarial training. Secondly, beyond the $\ell_2$ norm, we explore alternative loss functions that are better suited for differentiating the roles of primary and supportive training objectives. Thirdly, promoting diversity in the target probability distribution can further improve the model’s robustness.

\subsection{Adversarial mutual regularization (AMR)}
\label{Adversarial mutual}
\textit{Knowledge distillation (KD)} enhances the robustness of student models by enabling them to inherit soft target distributions from teacher models, typically through the use of KL-divergence. In addition, mutual learning leverages a pair of symmetric KL divergence terms to facilitate more effective knowledge exchange between models in distillation scenarios. However, although some studies adopt the squared \( \ell_2 \) norm to enforce consistency between adversarial outputs, which can improve robustness, the robustness achieved by these methods remains limited. Our work reflects on these limitations and leads to key findings and insights.

Inspired by the mathematical similarity between KL-divergence and MSE, we propose a role-aware regularization strategy that assigns different weights to a pair of symmetric KL-divergence terms—designating one as the primary objective and the other as auxiliary. This design explicitly distinguishes the dominant and supportive roles of each output in the adversarial training process, leading to more effective optimization and improved model robustness:
\begin{equation}
\begin{split}
     \mathcal{L}_\text{AMR}(x,y) = &\ \alpha\, \mathcal{L}_\text{KL}\left(f_{\boldsymbol{w}}(\boldsymbol{x}+\boldsymbol{\delta}^{\text{final}}) \parallel  f_{\boldsymbol{w}}(\boldsymbol{x}+\boldsymbol{\delta}^{\text{initial}})\right) \\
     &+ \beta\, \mathcal{L}_\text{KL}\left(f_{\boldsymbol{w}}(\boldsymbol{x}+\boldsymbol{\delta}^{\text{initial}}) \parallel  f_{\boldsymbol{w}}(\boldsymbol{x}+\boldsymbol{\delta}^{\text{final}})\right),
\end{split}
\label{eq:AMR_loss_function}
\end{equation}
where $\alpha$ and $\beta$ are weighting factors that control the trade-off between the two KL divergence terms during training. $\mathcal{L}_{\text{KL}}$ denotes the KL divergence loss. In the first term, $f_{\boldsymbol{w}}(\boldsymbol{x} + \boldsymbol{\delta}^{\text{initial}})$ serves as the target distribution, while in the second term, it acts as the predicted distribution. Conversely, the final adversarial output $f_{\boldsymbol{w}}(\boldsymbol{x} + \boldsymbol{\delta}^{\text{final}})$ serves as the predicted distribution in the first term and the target distribution in the second.

\subsection{Adversarial generalization regularization (AGR)}
\label{Adversarial Generalization}
Although the initialization of perturbation $\boldsymbol{\delta}^{\text{initial}}$ promotes smoother loss landscapes during adversarial training, the corresponding output $f_{\boldsymbol{w}}(\boldsymbol{x} + \boldsymbol{\delta}^{\text{initial}})$, when used as a target distribution, often lacks correct classification accuracy. We note that some methods aim to improve classification accuracy. For example, DGAD~\cite{park2024dynamic} proposed a strategy that replaces the teacher’s maximum prediction probability with that of a misclassified class, thereby aligning the teacher’s output with the ground-truth label. This technique has shown effectiveness in enhancing robustness in student-teacher frameworks. However, when we adopt and adjust either clean or adversarial probability distributions to ensure correct classification alignment in this work, we observe a decline in robust accuracy. We hypothesize that manually modifying the probability distribution in a fixed or predefined manner may lead to overconfident predictions, ultimately impairing robustness.

\begin{figure}[htbp]
    \centering
    \includegraphics[width=0.98\linewidth]{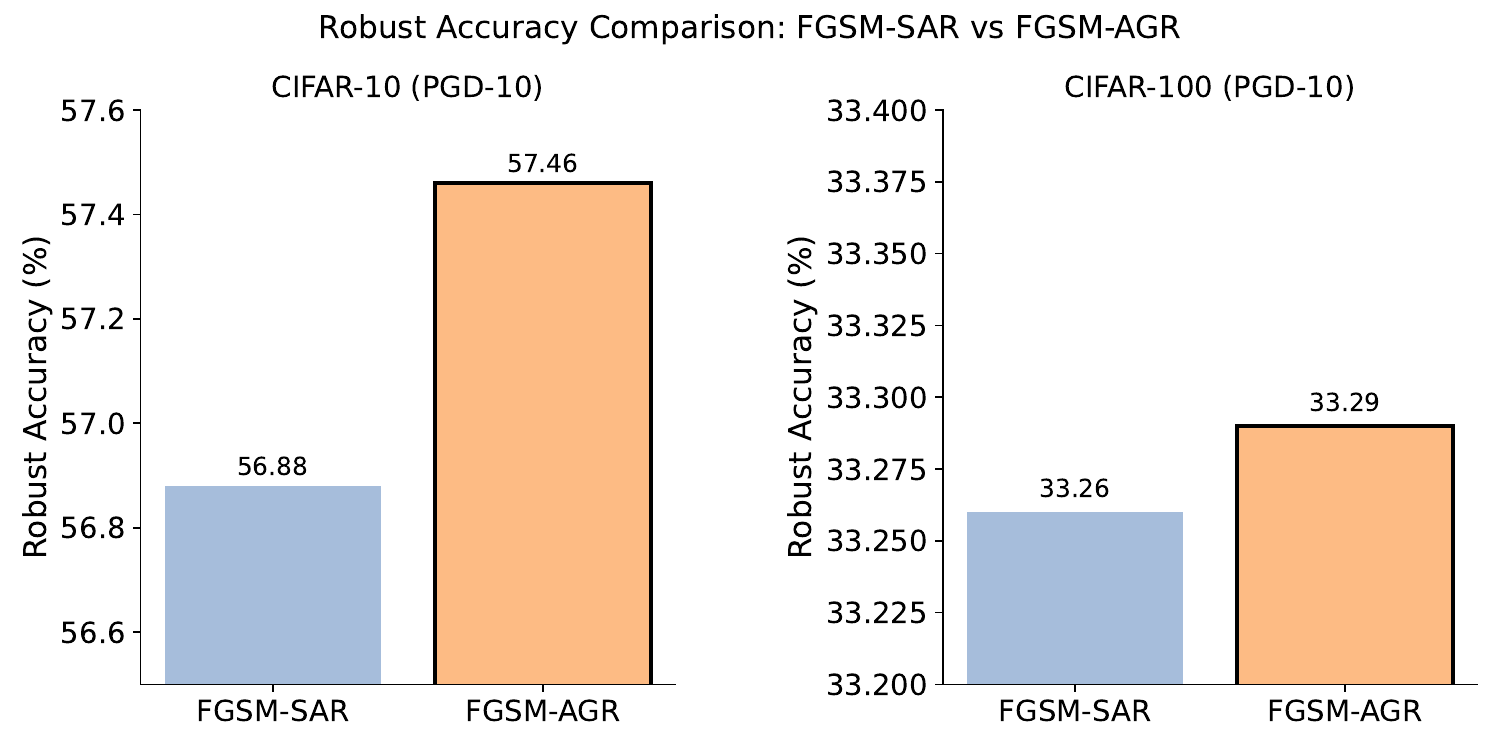}  
    \caption{Robust accuracy under PGD-10 attack on CIFAR-10 and CIFAR-100. FGSM-AGR (with the generalization term) consistently outperforms FGSM-SAR (without the generalization term) across both datasets.}

    \label{fig:AMR_AGR_comp}
\end{figure}

Building upon the above analysis, we observe that although the probability distribution of a model may achieve near-perfect accuracy on clean samples, this does not guarantee improved robustness against adversarial perturbations. Moreover, the adoption of initial perturbations may lead to model misclassification and degrade classification accuracy. Consequently, some studies~\cite{cui2021learnable} have suggested that clean output knowledge also plays a crucial role in enhancing model robustness. Therefore, we further introduce the incorporation of clean output knowledge without perturbation into adversarial training. In addition, we believe that effective adversarial training should incorporate both the knowledge from initial adversarial examples and clean samples. Specifically, we propose an approach that incorporates both initial adversarial input techniques and clean output, which can be viewed as enhancing the generalization of adversarial learning between the final adversarial output under adversarial training. In this approach, the gradients of the initial adversarial output, final adversarial output, and clean output are all involved in backpropagation: 
\begin{equation}
\begin{split}
\mathcal{L}_\text{AGR}(x,y) =\ 
    & \alpha\,   \mathcal{L}_\text{KL}\big(
        f_{\boldsymbol{w}}(\boldsymbol{x} + \boldsymbol{\delta}^{final})\ \big\|\ 
        f_{\boldsymbol{w}}(\boldsymbol{x} + \boldsymbol{\delta}^{initial})
    \big) \\
  +\ & \beta\,  \mathcal{L}_\text{KL}\big(
        f_{\boldsymbol{w}}(\boldsymbol{x} + \boldsymbol{\delta}^{initial})\ \big\|\ 
        f_{\boldsymbol{w}}(\boldsymbol{x} + \boldsymbol{\delta}^{final})
    \big) \\
  +\ & \gamma\,  \mathcal{L}_\text{KL}\big(
        f_{\boldsymbol{w}}(\boldsymbol{x} + \boldsymbol{\delta}^{final})\ \big\|\ 
        f_{\boldsymbol{w}}(\boldsymbol{x})
    \big),
\end{split}
\label{eq:AGR_loss_function}
\end{equation}
where $f_{\boldsymbol{w}}(\boldsymbol{x})$ denotes the clean output, which serves as an additional target for the final adversarial output $f_{\boldsymbol{w}}(\boldsymbol{x} + \boldsymbol{\delta}^{\text{final}})$, a strategy we refer to as \textit{Adversarial Generalization Regularization (AGR)}. In addition to enforcing consistency between the initial and final adversarial outputs, AGR further guides the final adversarial output toward the clean prediction. By increasing the weighting factor $\gamma$, the influence of the clean output in the optimization increases, thereby enhancing the model’s generalization and robustness under adversarial settings. Fig.~\ref{fig:AMR_AGR_comp} demonstrates that encouraging target output diversity, guided by clean examples during adversarial training, contributes to improved model robustness.




\section{Experiment and Results}
\label{sec:exp_results}
To evaluate the effectiveness of the proposed method, we perform comprehensive experiments on various network architectures and benchmark datasets, including CIFAR-10~\cite{krizhevsky2009learning}, CIFAR-100~\cite{krizhevsky2009learning}, and Tiny ImageNet~\cite{deng2009imagenet}. The proposed method is compared against several state-of-the-art fast adversarial training techniques, including FGSM-PGI~\cite{jia2022prior} and FGSM-PGK~\cite{jia2024improving}. Additionally, we assess the performance of the proposed method by comparing it with the robustness of the model under the same and even stronger perturbation settings.


\subsection{Experiment Settings}

For comparisons on CIFAR-10, we evaluate our method using both WideResNet-30-10 and ResNet-18. On CIFAR-100, consistent with the baseline setting using ResNet-18. For evaluations on Tiny-ImageNet, we employ ResNet-18 and compare with other fast adversarial training methods. We analyze robustness under varying perturbation magnitudes on CIFAR-10 by considering $\ell_\infty$ bounds $\epsilon \in \{8/255, 10/255, 12/255, 14/255, 16/255\}$. In addition, we investigate the contributions of the loss trade-off parameter $\alpha$ and $\beta$ in Eq.~\ref{eq:AMR_loss_function}, and $\gamma$ in Eq.~\ref{eq:AGR_loss_function}. An appropriate value of $\alpha$ is responsible for the main objective of adversarial training; therefore, it is typically set to be the largest among the weights. The term $\beta$ helps prevent overfitting during adversarial training and is usually set to about half of $\alpha$. The coefficient $\gamma$ controls the degree of generalization in adversarial training and is typically set to be comparable to $\beta$. Following FGSM-PGK as our baseline, all training hyperparameters on CIFAR-10 and CIFAR-100 (such as the learning rate, optimizer, and total number of training epochs) are kept consistent with the baseline settings to ensure a fair comparison.

\subsection{Results}
\subsubsection{Results on CIFAR-10}
\begin{table*}[t]
\centering
\caption{Clean and robust accuracy (\%) of ResNet-18 on the CIFAR-10 dataset under $\ell_{\infty} = 8/255$ perturbations, evaluated at both the best and last checkpoints. The best results are highlighted in bold.}
\label{table:cifar10}
\setlength\tabcolsep{14pt}
\renewcommand{\arraystretch}{1.2}
\begin{tabular}{lcccccc}
\toprule
Method & \begin{tabular}[c]{@{}c@{}}Clean\\ Best/Last\end{tabular} 
       & \begin{tabular}[c]{@{}c@{}}PGD-10\\ Best/Last\end{tabular} 
       & \begin{tabular}[c]{@{}c@{}}PGD-20\\ Best/Last\end{tabular} 
       & \begin{tabular}[c]{@{}c@{}}PGD-50\\ Best/Last\end{tabular} 
       & \begin{tabular}[c]{@{}c@{}}C\&W\\ Best/Last\end{tabular} 
       & \begin{tabular}[c]{@{}c@{}}AA\\ Best/Last\end{tabular} \\
\midrule
PGD-AT & 82.65/82.32 & 53.39/53.76 & 52.52/52.83 & 52.27/52.60 & 51.28/51.08 & 48.93/48.68 \\
\midrule
FGSM-RS~\cite{wong2020fast} & 73.81/83.82 & 42.31/0.09 & 41.55/0.04 & 41.26/0.02 & 39.84/0.00 & 37.07/0.00 \\
FGSM-CKPT~\cite{kim2021understanding} & \textbf{90.29/90.29} & 41.96/41.96 & 39.84/39.84 & 39.15/39.15 & 41.13/41.13 & 37.15/37.15 \\
FGSM-SDI~\cite{jia2022boosting} & 84.86/85.25 & 53.73/53.18 & 52.54/52.05 & 52.18/51.79 & 51.00/50.29 & 48.50/47.91 \\
NuAT~\cite{sriramanan2021towards} & 81.58/81.38 & 53.96/53.52 & 52.90/52.65 & 52.61/52.48 & \textbf{51.30}/50.63 & 49.09/48.70 \\
GAT~\cite{sriramanan2020guided} & 79.79/80.41 & 54.18/53.29 & 53.55/52.06 & 53.42/51.76 & 49.04/49.07 & 47.53/46.56 \\
FGSM-GA~\cite{andriushchenko2020understanding} & 83.96/84.43 & 49.23/48.67 & 47.57/46.66 & 46.89/46.08 & 47.46/46.75 & 43.45/42.63 \\
Free-AT (m=8)~\cite{shafahi2019adversarial} & 80.38/80.75 & 47.10/45.82 & 45.85/44.82 & 45.62/44.48 & 44.42/43.73 & 42.17/41.17 \\
FGSM-PGI~\cite{jia2022prior} & 81.72/81.72 & 55.18/55.18 & 54.36/54.36 & 54.17/54.17 & 50.75/50.75 & 49.00/49.00 \\
FGSM-PGK~\cite{jia2024improving} & 81.58/81.58 & 56.08/56.08 & 55.51/55.51 & 55.31/55.31 & 51.17/51.17 & 49.51/49.51 \\
\rowcolor{LightCyan}FGSM-AMR (ours) & 81.43/81.43 & 56.88/56.88 & 56.35/56.35 & 56.13/56.13 & 51.32/51.32 & 49.73/49.73 \\
\rowcolor{LightCyan}FGSM-AGR (ours) & 80.08/80.28 & \textbf{57.46}/57.44 &  56.91/\textbf{56.95} & \textbf{56.83/56.83} & 51.73/\textbf{51.74} & \textbf{49.86}/49.84 \\
\bottomrule
\end{tabular}
\end{table*}

We first evaluate our method on the CIFAR-10 dataset using the ResNet-18 model and compare it against existing approaches. As shown in Table~\ref{table:cifar10}, under PGD-10 attacks, our method achieves a robust accuracy of 57.46$\%$, outperforming FGSM-PGK by 1.38$\%$. Furthermore, as illustrated in Fig.~\ref{fig:var-eps-robustness}, our method not only achieves higher robustness under the same perturbation settings, but also consistently outperforms FGSM-PGI and FGSM-PGK under stronger $\ell_\infty$ attacks, including PGD and AA. Under a more challenging perturbation level of $\ell_\infty = 16/255$, our method attains 49.42\% robust accuracy, exceeding FGSM-PGK by 1.25$\%$.

\begin{table}
\centering
\caption{Clean and robust accuracy (\%) of WideResNet-34-10 on CIFAR-10 under $\ell_{\infty} = 8/255$ and $16/255$ perturbations. Both best and last checkpoints are reported. Best results are in bold.}
\label{table:cifar10-Wide-combined}
\setlength\tabcolsep{0.05cm}
\renewcommand{\arraystretch}{1.5}
\begin{tabular}{lccccccc}
\toprule
Method & Epoch & Clean & PGD-10 & PGD-20 & PGD-50 & C\&W & AA \\
\midrule
\multicolumn{8}{c}{$\ell_\infty = 8/255$} \\
\midrule
FGSM-PGK~\cite{jia2024improving} & Best & 81.27 & 59.92 & 59.26 & 59.18 & 52.36 & 52.36 \\
                                  & Last & 66.07 & 45.40 & 45.22 & 45.18 & 40.92 & 40.25 \\
                                  
\rowcolor{LightCyan}FGSM-AMR (ours)                  & Best & 84.43 & 61.28 & 60.47 & 60.26 & 56.01 & 52.79 \\
\rowcolor{LightCyan}                                  & Last & \textbf{84.85} & 61.12 & 60.32 & 60.19 & 56.18 & 52.85 \\

\rowcolor{LightCyan}FGSM-AGR (ours)                  & Best & 83.43 & \textbf{61.10} & \textbf{60.71} & \textbf{60.60} & \textbf{55.82} & \textbf{53.04} \\
\rowcolor{LightCyan}                                  & Last & 83.43 & \textbf{61.10} & \textbf{60.71} & \textbf{60.60} & \textbf{55.82} & \textbf{53.04} \\
\midrule
\multicolumn{8}{c}{$\ell_\infty = 16/255$} \\
\midrule
FGSM-PGK~\cite{jia2024improving} & Best & - & 48.17 & 44.78 & 43.80 & - & 34.81 \\
                                  & Last & - & 36.01 & 34.35 & 34.04 & - & 26.67 \\ 
\rowcolor{LightCyan}FGSM-AMR (ours)                  & Best & - & 48.23 & 44.70 & 43.55 & - & 33.70 \\
\rowcolor{LightCyan}                                  & Last & - & 47.92 & 44.02 & 42.96 & - & 33.25 \\
\rowcolor{LightCyan}FGSM-AGR (ours)                  & Best & - & \textbf{49.42} & \textbf{45.83} & \textbf{44.71} & - & \textbf{34.59} \\
\rowcolor{LightCyan}                                 & Last & - & 49.21 & 45.42 & 44.55 & - & 34.32 \\
\bottomrule
\end{tabular}
\end{table}

Furthermore, in terms of training efficiency, although the KL divergence loss is computationally slower than the MSE loss, its superior effectiveness enables our method to achieve better performance with only 95 training epochs. In contrast, FGSM-PGI and FGSM-PGK require 110 epochs to reach comparable performance, indicating that our approach converges faster. On the CIFAR-10 dataset, due to differences in hardware configurations, we compare our results only with the most recent and state-of-the-art methods under the setting of 98 training epochs.

\begin{figure}[htbp]
    \centering
    \includegraphics[width=\linewidth]{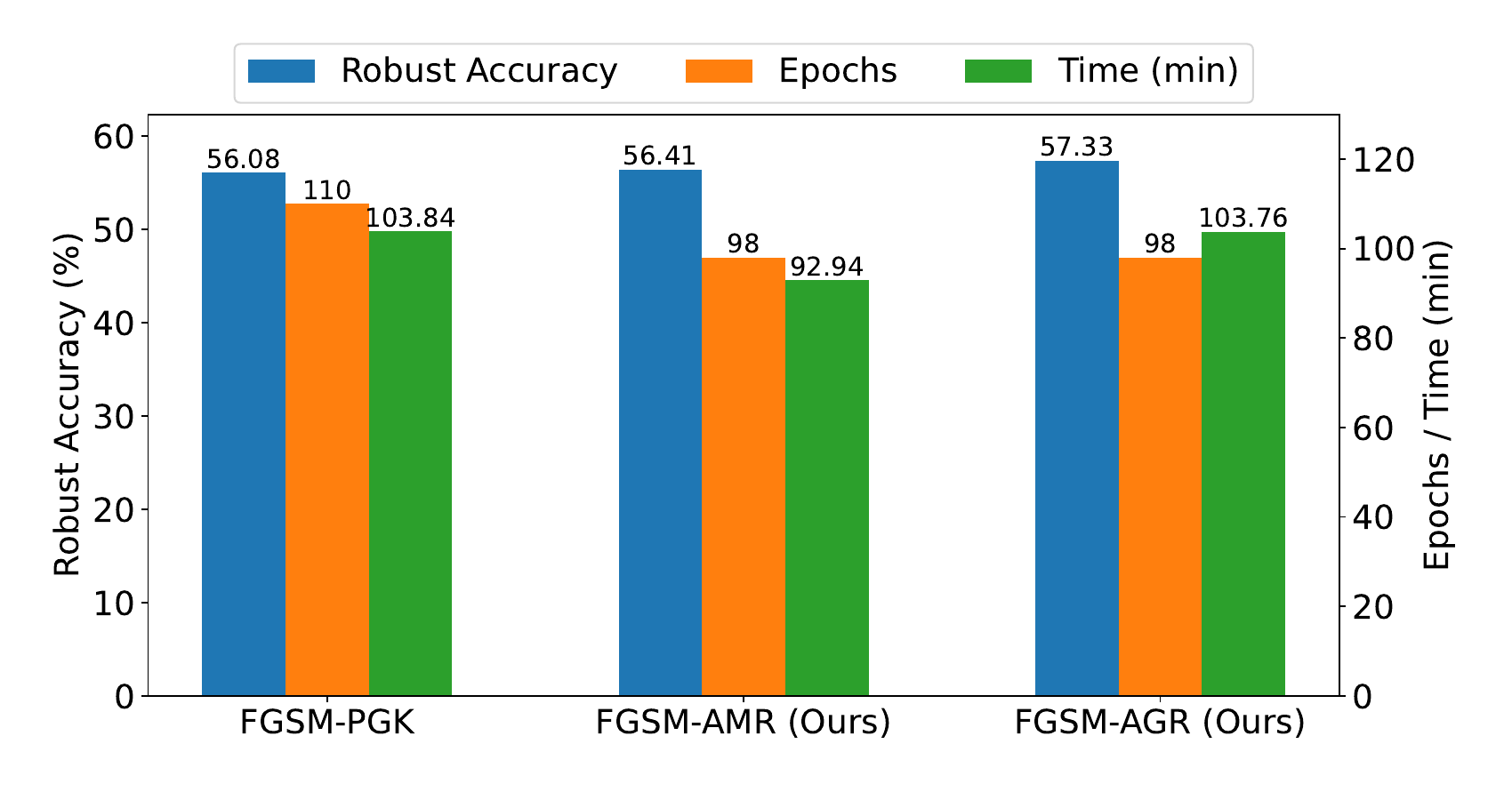}
    \caption{
        Comparison between our methods (FGSM-AMR / FGSM-AGR) and the SOTA baseline (FGSM-PGK) on CIFAR-10 under PGD-10 attack. 
        This plot shows both robustness (PGD-10 attack) and training cost (epochs and time in minutes). Our experiments were conducted on a single NVIDIA RTX 3090 Ti GPU.
    }
    \label{fig:fgsm_sota_comparison}
\end{figure}


We further validate the effectiveness of our approach using the WideResNet-34-10 architecture. As shown in Tables~\ref{table:cifar10-Wide-combined} and Fig.~\ref{fig:var-eps-robustness}, our method consistently achieves the highest robust accuracy across all evaluated attack settings compared to existing methods. Notably, it not only maintains strong robustness under the same perturbation budgets, but also performs better under stronger $\ell_\infty$ attacks. Specifically, under PGD-10 attacks, our method achieves a robust accuracy of 57.46\%, outperforming FGSM-PGK by 1.38\%. When the perturbation is increased to $\ell_\infty = 16/255$, our method still achieves 38.48\% robust accuracy, which is 1.86\% higher than FGSM-PGK.

\subsubsection{Results on CIFAR-100}

We evaluate the effectiveness of our proposed adversarial training method on the CIFAR-100 dataset using the ResNet-18 architecture. We compare our approach with the previous fast adversarial training method. The experimental results, summarized in Table~\ref{table:cifar100}, demonstrate that our ResNet-18 models trained with the FGSM-AMR (Adversarial Mutual Regularization) loss and AGR (Adversarial Generalization Regularization) loss exhibit superior robustness compared to the baseline. Among them, FGSM-AGR achieves the highest robustness. Under PGD-50 attacks, it achieves a robust accuracy of 33.30\%, outperforming FGSM-PGK by 0.47\%. Under the AutoAttack (AA), FGSM-AGR reaches 27.42\%, which is 0.56\% higher than FGSM-PGK.

\begin{table*}[t]
\centering
\caption{Clean and robust accuracy (\%) of ResNet-18 on the CIFAR-100 dataset under $\ell_{\infty} = 8/255$ perturbations, evaluated at both the best and last checkpoints. The best results are highlighted in bold.}
\label{table:cifar100}
\setlength\tabcolsep{12pt}
\renewcommand{\arraystretch}{1.2}
\begin{tabular}{lcccccc}
\toprule
Method & \begin{tabular}[c]{@{}c@{}}Clean\\ Best/Last\end{tabular} 
       & \begin{tabular}[c]{@{}c@{}}PGD-10\\ Best/Last\end{tabular} 
       & \begin{tabular}[c]{@{}c@{}}PGD-20\\ Best/Last\end{tabular} 
       & \begin{tabular}[c]{@{}c@{}}PGD-50\\ Best/Last\end{tabular} 
       & \begin{tabular}[c]{@{}c@{}}C\&W\\ Best/Last\end{tabular} 
       & \begin{tabular}[c]{@{}c@{}}AA\\ Best/Last\end{tabular} \\
\midrule
PGD-AT~\cite{rice2020overfitting}           & 57.52/57.50 & 29.60/29.54 & 28.99/29.00 & 28.87/28.90 & 28.85/27.60 & 25.48/25.58 \\
\midrule
FGSM-RS~\cite{wong2020fast}                 & 49.85/60.55 & 22.47/0.45  & 22.01/0.25  & 21.82/0.19  & 20.55/0.25  & 18.29/0.00  \\
FGSM-CKPT~\cite{kim2021understanding}       & \textbf{ 60.93/60.93 } & 16.58/16.69 & 15.47/15.61 & 15.19/15.24 & 16.40/16.60 & 14.17/14.34 \\
FGSM-SDI~\cite{jia2022boosting}             & 60.67/60.82 & 31.50/30.87 & 30.89/30.34 & 30.60/30.08 & 27.15/27.30 & 25.23/25.19 \\
NuAT~\cite{sriramanan2021towards}          & 59.71/59.62 & 27.54/27.07 & 23.02/22.72 & 20.18/20.09 & 22.07/21.59 & 11.32/11.55 \\
GAT~\cite{sriramanan2020guided}            & 57.01/56.07 & 24.55/23.92 & 23.80/23.18 & 23.55/23.00 & 22.02/21.93 & 19.60/19.51 \\
FGSM-GA~\cite{andriushchenko2020understanding} & 54.35/55.10 & 22.93/20.04 & 22.36/19.13 & 22.20/18.84 & 21.20/18.96 & 18.88/16.45 \\
Free-AT (m=8)~\cite{shafahi2019adversarial} & 52.49/52.63 & 24.07/22.86 & 23.52/22.32 & 23.36/22.16 & 21.66/20.68 & 19.47/18.57 \\
FGSM-PGI~\cite{jia2022prior}               & 58.78/58.81 & 31.78/31.60 & 31.26/31.06 & 31.14/30.88 & 28.06/27.72 & 25.67/25.42 \\
FGSM-PGK~\cite{jia2024improving}                                      & 56.27/58.13 & 33.15/32.38 & 32.85/31.90 & 32.83/31.87 & 28.39/27.95 & 26.86/26.35 \\
\rowcolor{LightCyan}FGSM-SAR (ours)                               & 56.08/55.71 & 33.26/33.06 & 32.93/32.86 & 32.84/32.68 & 28.64/28.89 & 27.27/27.22 \\
\rowcolor{LightCyan}FGSM-AGR (ours)                            & 53.57/53.57 & \textbf{ 33.29/33.29 } & \textbf{ 33.02/33.02 } & \textbf{ 32.95/32.95 } & \textbf{ 28.91/28.91 } & \textbf{ 27.42/27.42 } \\
\bottomrule
\end{tabular}
\end{table*}

\subsubsection{Results on Tiny ImageNet}

On the Tiny ImageNet dataset, we use images resized to 32$\times$32 resolution for training. Based on the ResNet-18 architecture, our proposed FGSM-AGR method achieves improved robustness. The training settings include a reset epoch of 10, momentum of 0.9, factor of 0.5, $\lambda$ = 1, and an EMA value of 0.8. Our method demonstrates superior robustness, achieving a robust accuracy of 13.79\% under the PGD-10 attack.

\begin{table}
\centering
\caption{Comparisons of clean and robust accuracy (\%) on the Tiny ImageNet database using ResNet18 with different adversarial training methods under $\ell_{\infty}=8/225$. The number in bold indicates the best result.}
\label{table:Tiny_ImageNet}
\setlength\tabcolsep{2pt}
\renewcommand{\arraystretch}{1.2}
\begin{tabular}{lccccc}
\toprule
Method       & \begin{tabular}[c]{@{}c@{}}Clean\\Best/Last\end{tabular} 
             & \begin{tabular}[c]{@{}c@{}}PGD-10\\Best/Last\end{tabular} 
             & \begin{tabular}[c]{@{}c@{}}PGD-50\\Best/Last\end{tabular} 
             & \begin{tabular}[c]{@{}c@{}}C\&W\\Best/Last\end{tabular} 
             & \begin{tabular}[c]{@{}c@{}}AA\\Best/Last\end{tabular} \\
\midrule

FGSM-PGK~\cite{jia2024improving}                      &  \textbf{26.79}   & 13.66  & 13.55 & 10.32 & 9.20 \\
\rowcolor{LightCyan}FGSM-AGR (ours)                            & 26.42 & \textbf{13.79}  & \textbf{13.56} & \textbf{10.60} & \textbf{9.31} \\

\bottomrule
\end{tabular}
\end{table}



\subsection{Performance Analysis}




\begin{figure}[t]
    \centering
    \begin{subfigure}[t]{0.98\linewidth}
        \centering
        \includegraphics[width=\linewidth]{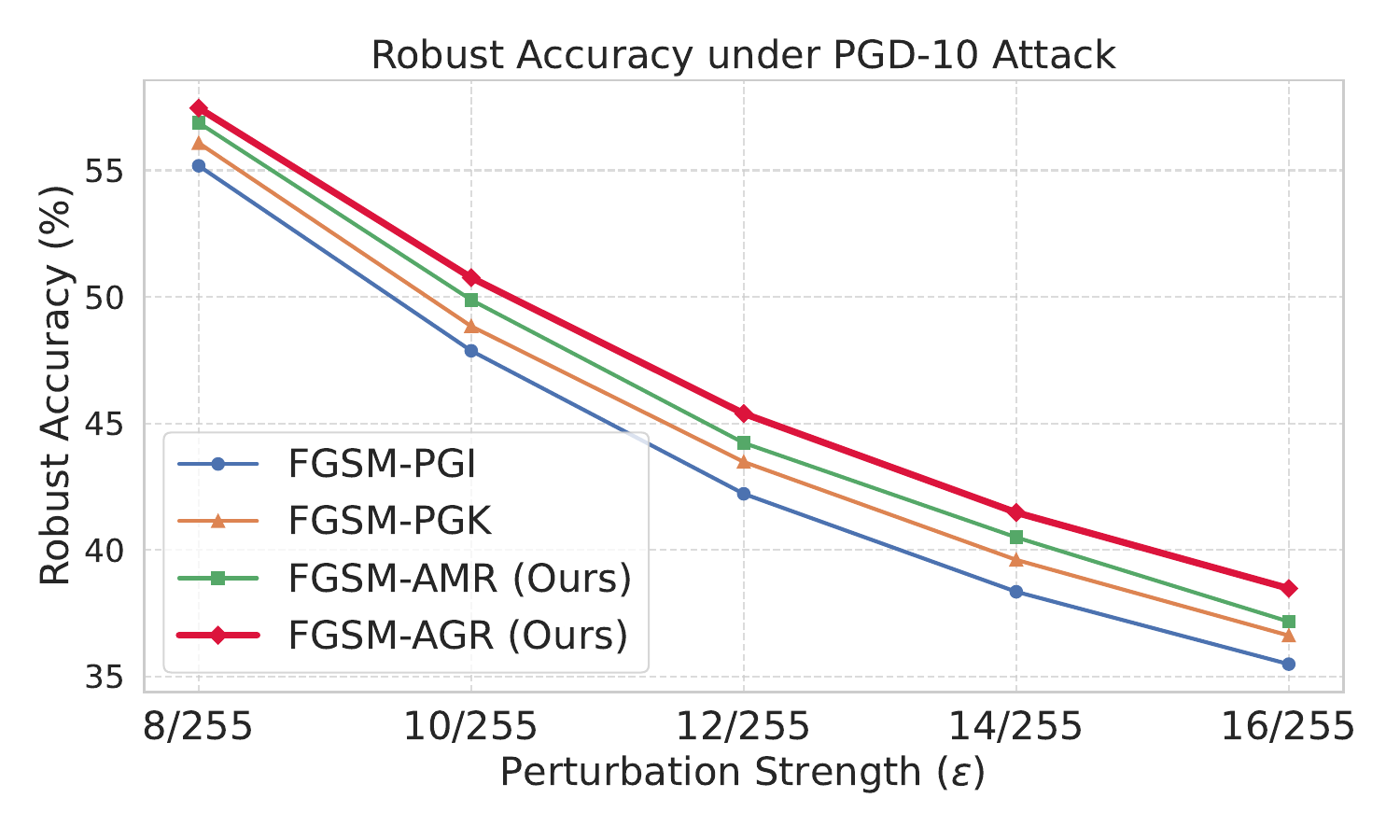}
        \caption{PGD-10 attack results.}
        \label{fig:pgd-eps}
    \end{subfigure}
    \hfill
    \begin{subfigure}[t]{0.98\linewidth}
        \centering
        \includegraphics[width=\linewidth]{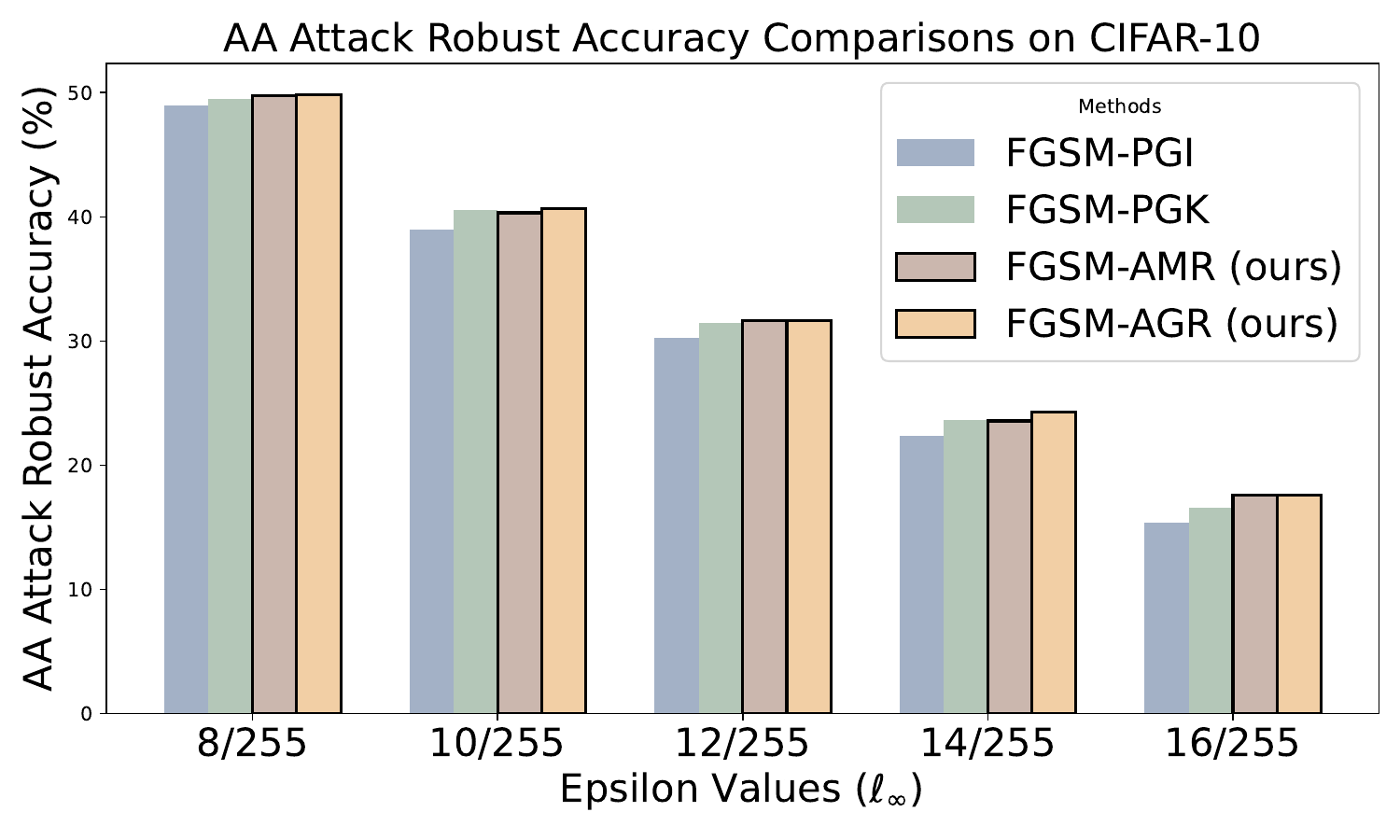}
        \caption{AutoAttack results.}
        \label{fig:aa-eps}
    \end{subfigure}
    \caption{Robustness evaluation on CIFAR-10 under varying $\ell_\infty$ perturbation bounds using ResNet-18. (a) Results under PGD-10 attacks. (b) Robust accuracy under AutoAttack for different PGD-2-AT based defenses. Our methods show consistent superiority robustness under stronger perturbations.}
    \label{fig:var-eps-robustness}
\end{figure}




\begin{figure}[h]
    \centering
    \begin{minipage}[t]{0.48\linewidth}
        \centering
        \includegraphics[width=\linewidth]{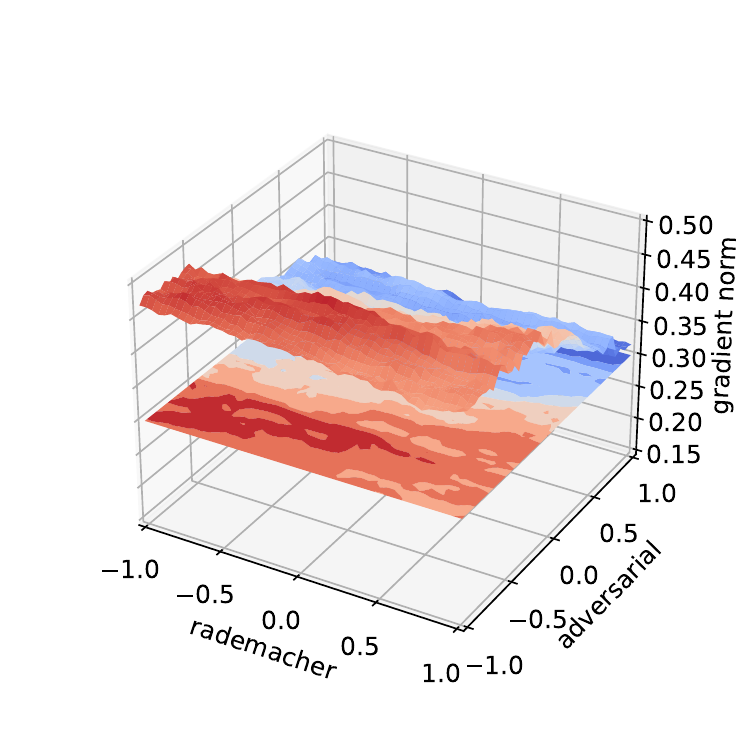}
        \vspace{-5mm}
        \subcaption{FGSM-PGI}
    \end{minipage}
    \hfill
    \begin{minipage}[t]{0.48\linewidth}
        \centering
        \includegraphics[width=\linewidth]{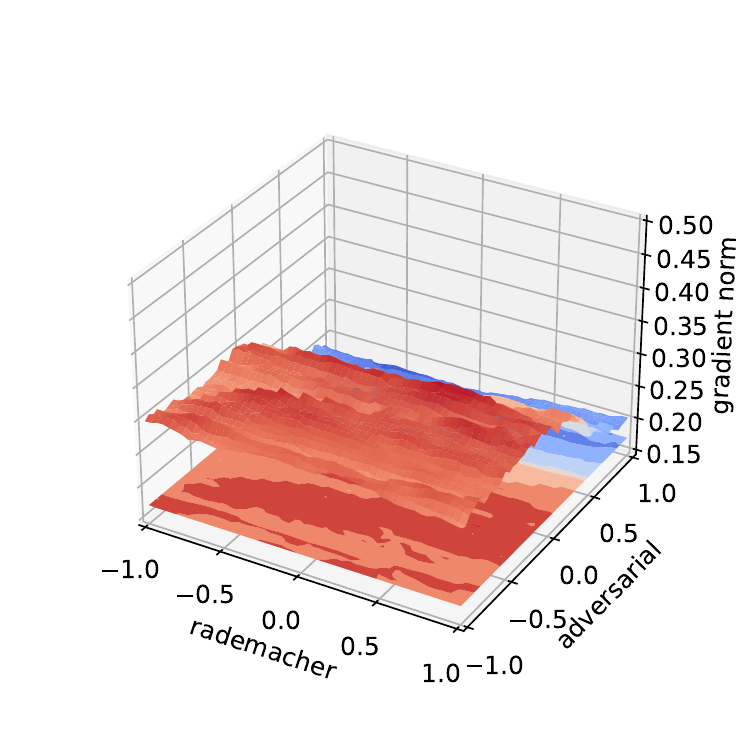}
        \vspace{-5mm}
        \subcaption{FGSM-PGK}
    \end{minipage}
    \begin{minipage}[t]{0.48\linewidth}
        \centering
        \includegraphics[width=\linewidth]{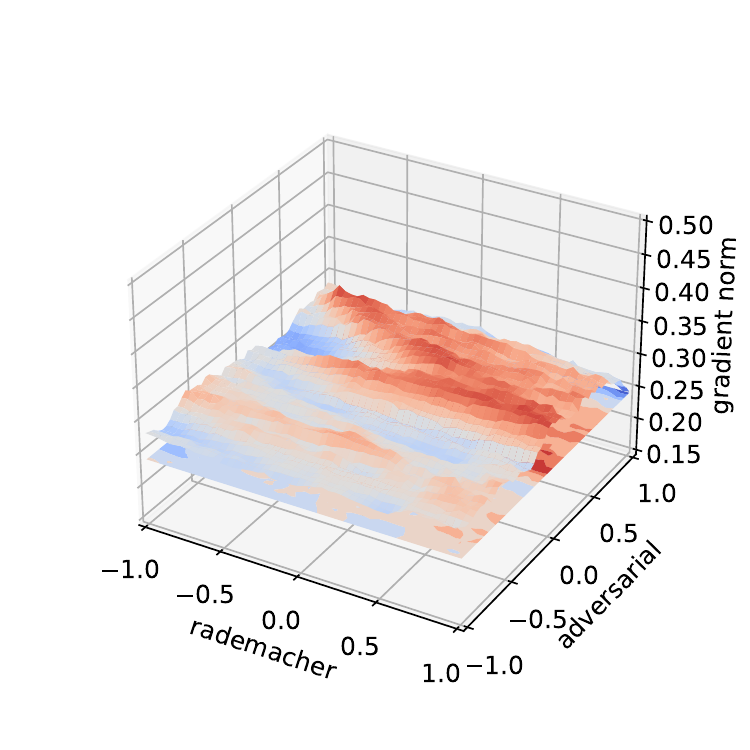}
        \vspace{-5mm}
        \subcaption{FGSM-AMR (ours)}
    \end{minipage}
    \hfill
    \begin{minipage}[t]{0.48\linewidth}
        \centering
        \includegraphics[width=\linewidth]{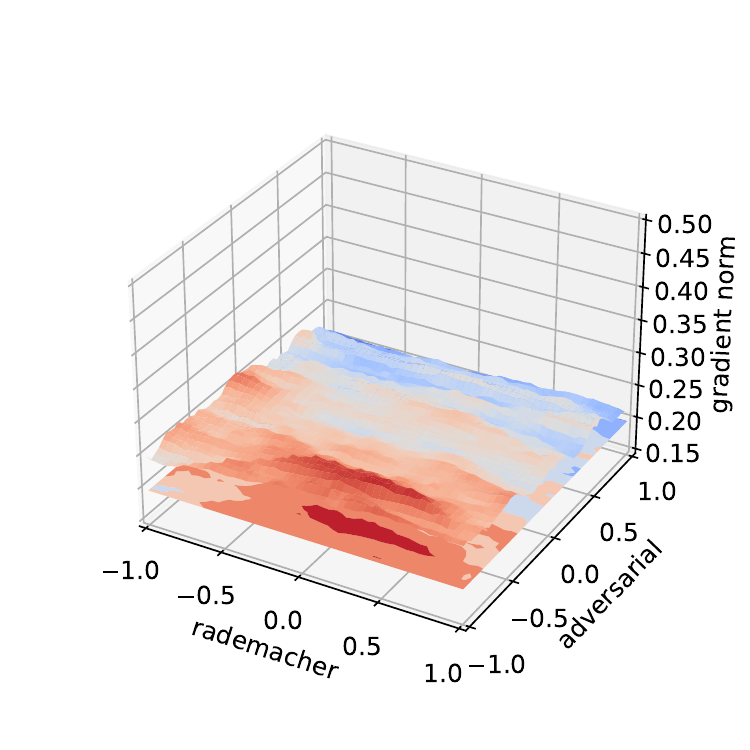}
        \vspace{-5mm}
        \subcaption{FGSM-AGR (ours)}
    \end{minipage}

    \vspace{-1mm} 
    \caption{
        Visualization of the loss landscapes for various PGD-2-AT adversarial training methods: (a) FGSM-PGI, (b) FGSM-PGK, (c) our proposed FGSM-AMR method, (d) our proposed FGSM-AGR method. Each subplot displays the cross-entropy loss over 1000 randomly selected CIFAR-10 test images, projected onto a 2D plane defined by a Rademacher direction \( v \sim \operatorname{Rademacher}(\eta) \) and an adversarial direction \( u = \eta \operatorname{sign}(\nabla_{x} f(\hat{x})) \) with \( \eta = 8/255 \), where the adversarial direction is obtained via PGD-10.
    }
    \label{fig:loss_landscape}
\end{figure}

To further evaluate the robustness of our method beyond attacks with the same perturbation strength, we also test it under stronger adversarial perturbations.Following the prior works FGSM-PGI~\cite{jia2022prior} and FGSM-PGK~\cite{jia2024improving}, we conduct experiments on the CIFAR-10 dataset using ResNet-18, and evaluate model performance under progressively stronger $\ell_\infty$ adversarial attack strengths. As shown in Fig.~\ref{fig:var-eps-robustness}, our method consistently outperforms baselines under stronger perturbations. Specifically, the results demonstrate that both \textit{Adversarial Mutual Regularization} and \textit{Adversarial Generalization Regularization} not only improve robustness against standard perturbations but also significantly enhance defense under stronger adversarial attacks.





\begin{figure*}[h!]
    \centering
    \begin{subfigure}{0.45\textwidth}
        \centering
        \includegraphics[width=\linewidth]{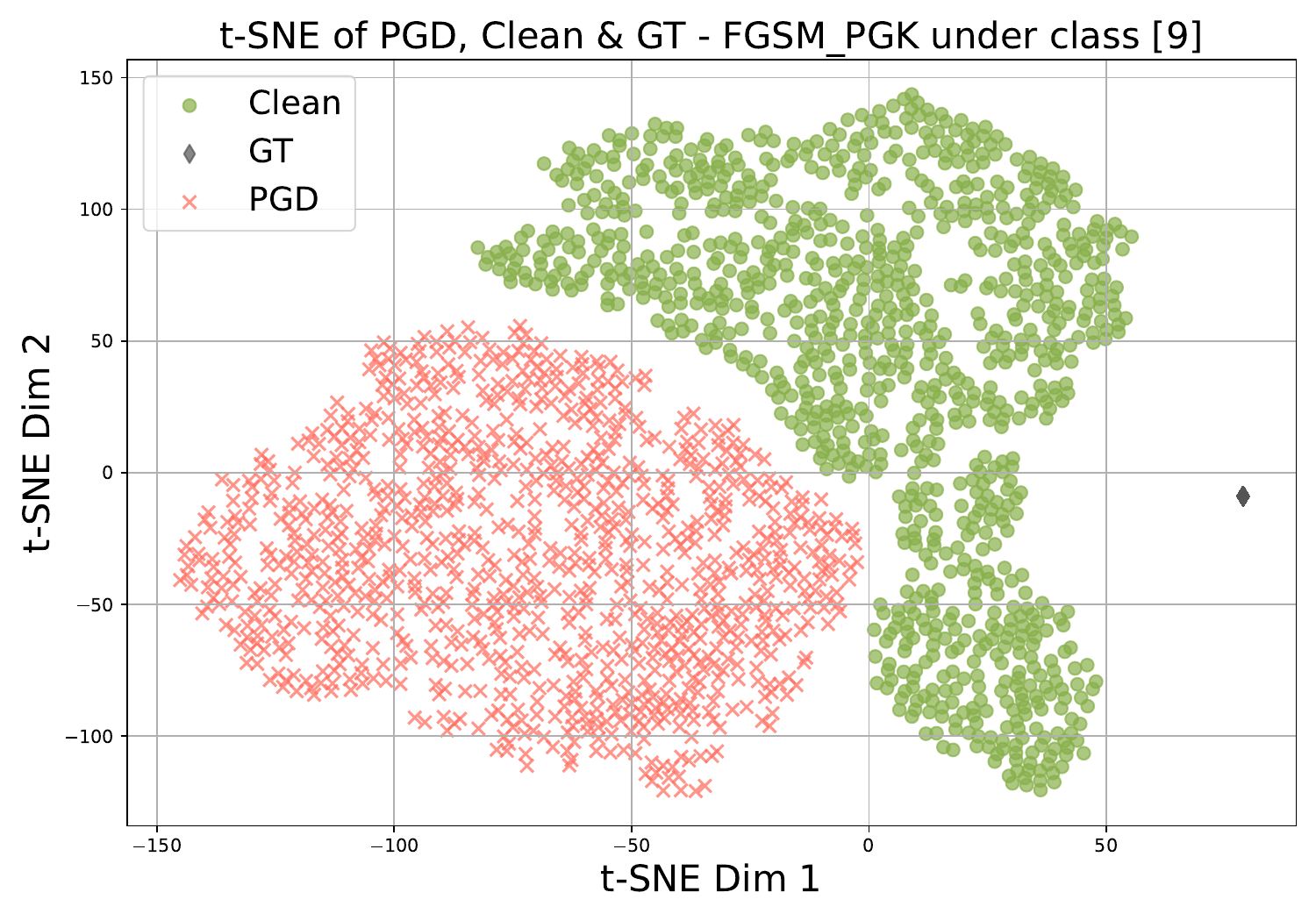}
        \caption{}
    \end{subfigure}
    \hfill
    \begin{subfigure}{0.45\textwidth}
        \centering
        \includegraphics[width=\linewidth]{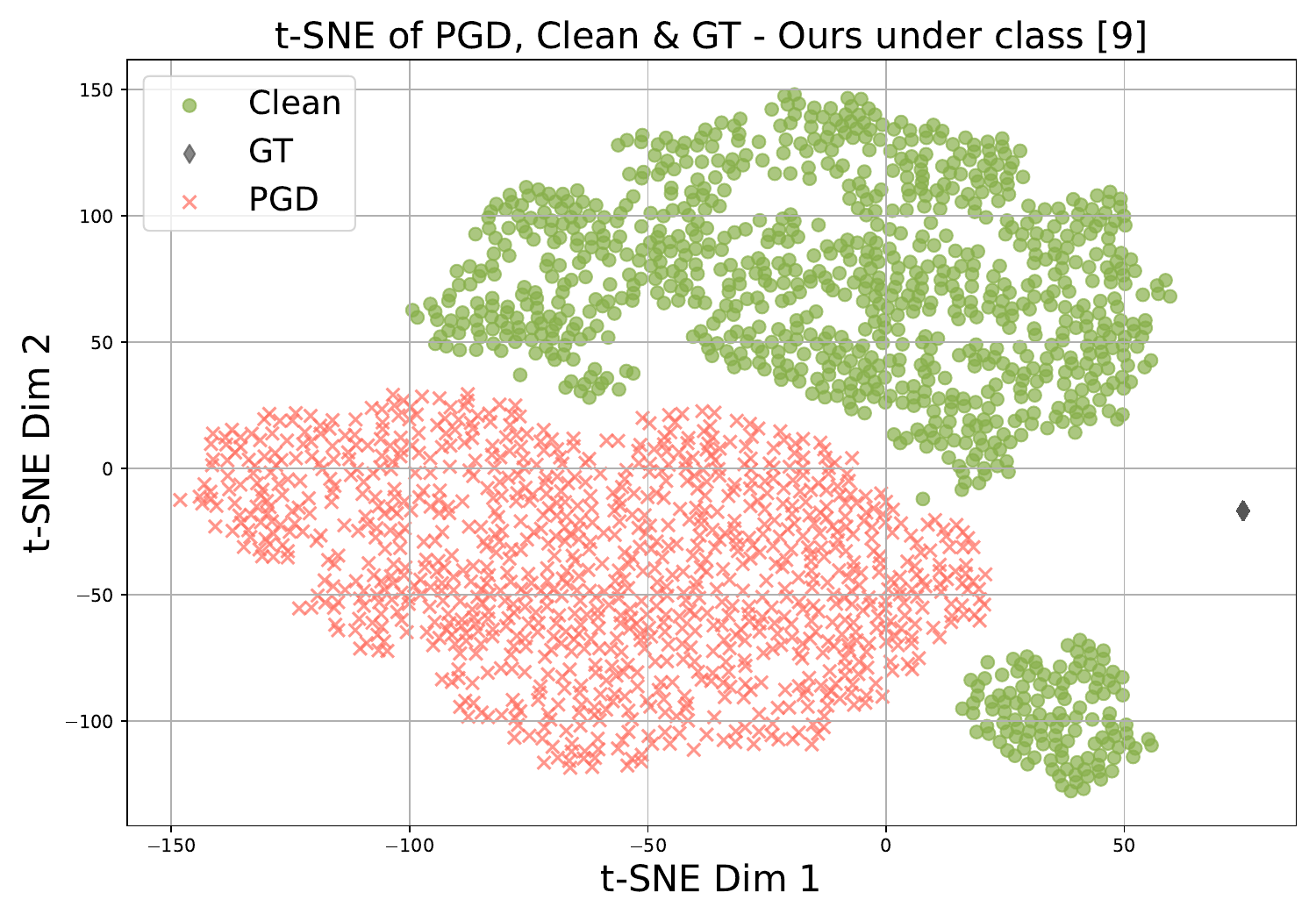}
        \caption{}
    \end{subfigure}
    \vskip\baselineskip
    \begin{subfigure}{0.45\textwidth}
        \centering
        \includegraphics[width=\linewidth]{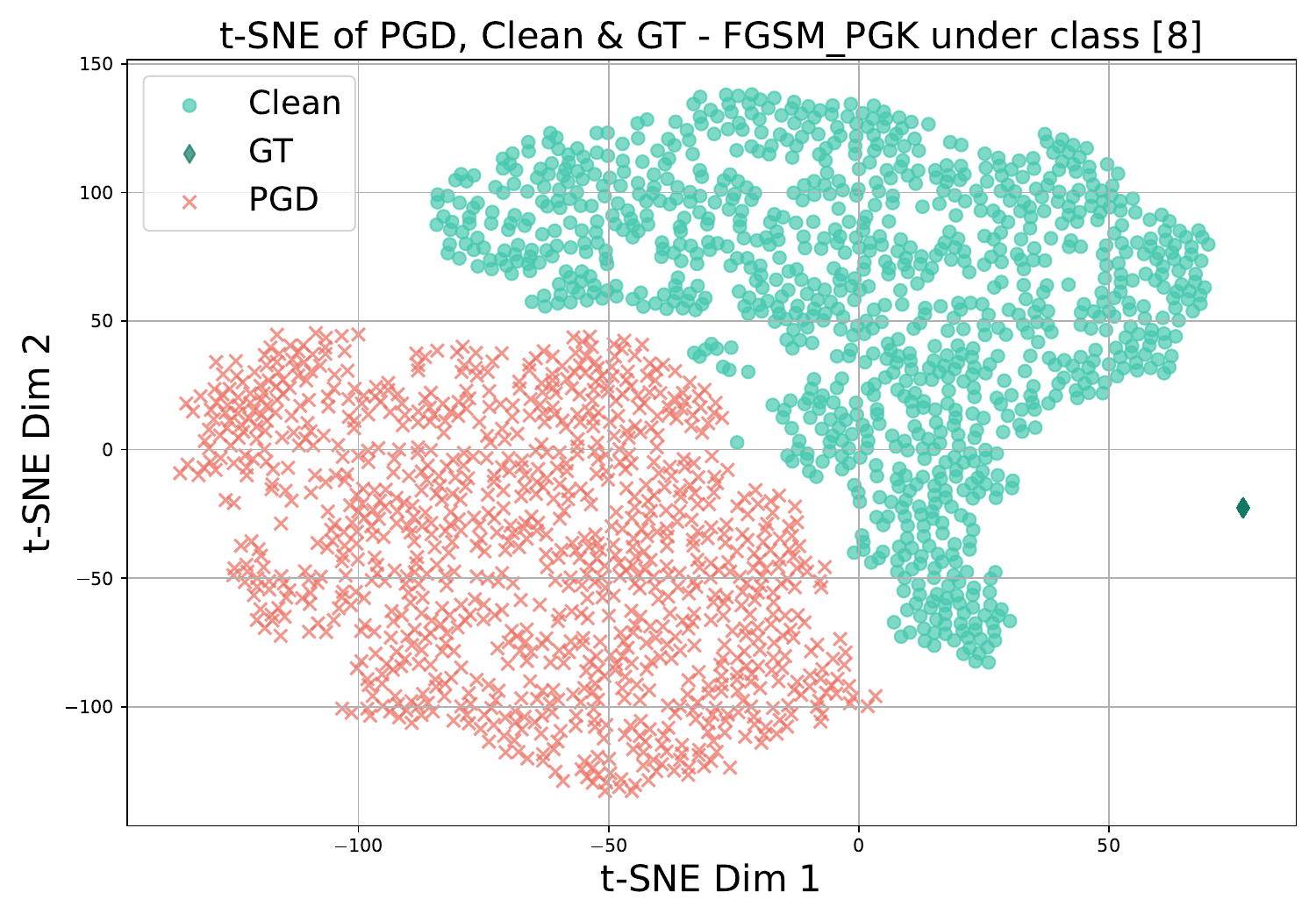}
        \caption{}
    \end{subfigure}
    \hfill
    \begin{subfigure}{0.45\textwidth}
        \centering
        \includegraphics[width=\linewidth]{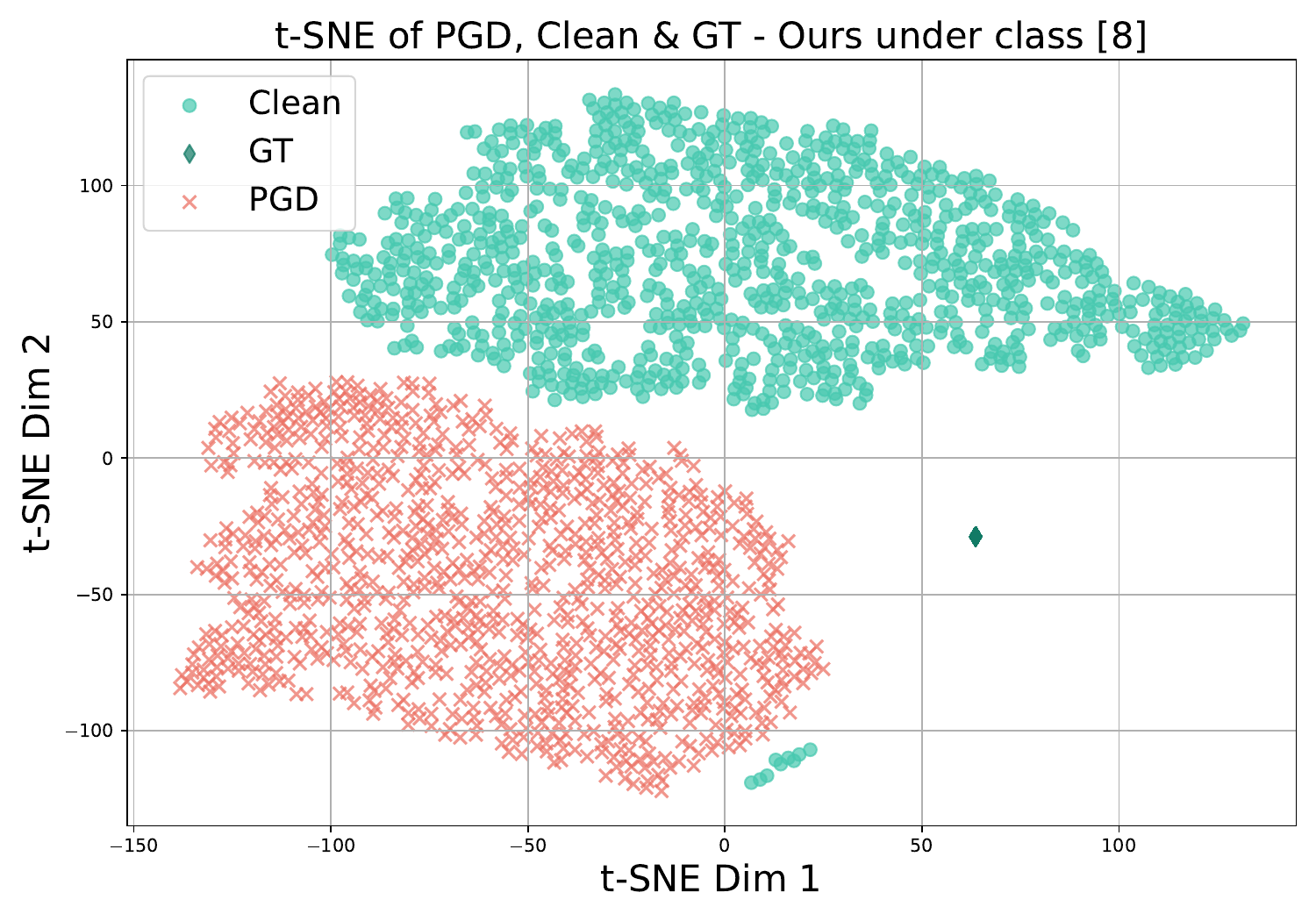}
        \caption{}
    \end{subfigure}
    \caption{
        t-SNE visualization of predicted adversarial and clean probability distributions, along with ground truth, for two classes (class 8 and class 9) using pre-trained models based on FGSM-PGK and our proposed method FGSM-AGR.
        The distribution produced by our method shows adversarial probabilities more concentrated around the ground truth class, indicating improved confidence and robustness.
    }
    \label{fig:probability_distributions}
\end{figure*}

To verify the effectiveness of our proposed method, we utilize visualization tools to compare it with existing approaches. Fig.~\ref{fig:loss_landscape} shows the loss landscapes of our method and previous PGD-2-AT based adversarial training techniques. In the colormap, red indicates high loss, blue indicates low loss, and white or lighter shades represent intermediate values. Compared to baselines, our method exhibits a smoother and flatter loss landscape, suggesting improved robustness and generalization. Fig.~\ref{fig:probability_distributions} visualizes the distributions of clean probabilities, adversarial probabilities, and ground truth labels. Our approach produces adversarial probability distributions closer to the ground truth in the t-SNE embedded space, demonstrating better alignment and enhanced robustness. We can see that in row 1, our method, compared to FGSM PGK, has ground truth separating adversarial probability distribution (red) from clean probability distribution much effectively (see ground truth appearing between green and red probabilities). Similar results were obtained in row two of Fig.~\ref{fig:probability_distributions}, where ground truth appears in between read and cyan points representing PGD-attack and clean probability distributions.



\subsection{Hyper-Parameter settings }
We investigate the contributions of the loss trade-off parameter $\alpha$ and $\beta$ in Eq.~\ref{eq:AMR_loss_function}, and $\gamma$ in Eq.~\ref{eq:AGR_loss_function}. An appropriate value of $\alpha$ is responsible for the main objective of adversarial training; therefore, it is typically set to be the largest among the weights. The term $\beta$ helps prevent overfitting during adversarial training and is usually set to about half of $\alpha$. The coefficient $\gamma$ controls the degree of generalization in adversarial training and is typically set to be comparable to $\beta$.

The effectiveness of our methods depends on appropriate hyperparameter configurations. For the CIFAR-10 dataset, we set the hyperparameters as $\alpha=50$, $\beta=20$, and $\gamma=20$, whereas for CIFAR-100, the values are set as $\alpha=1200$, $\beta=700$, and $\gamma=700$. These tailored settings enable FGSM-AMR and FGSM-AGR to consistently achieve stable and robust performance, contributing to improved model robustness. For the Tiny-ImageNet dataset, we set the FGSM-AGR hyperparameters to $\alpha = 200$, $\beta = 100$, and $\gamma = 100$.

\section{Conclusion}
\label{sec:con}
We propose a methodology called RegMix (adversarial mutual and generalization regularization) for enhancing deep neural networks (DNN) robustness. RegMix consists of two regularization strategies for the single-model setting: (i) \textit{mutual adversarial regularization},  which incorporates mutual learning from knowledge distillation into adversarial training; and (ii) \textit{adversarial generalization regularization}, which applies multi-objective adversarial training to improve generalization. Both methods significantly enhance model robustness, as shown by experiments on three datasets under different perturbation budgets, validating the effectiveness of our approach. We discover that training with probability distributions not only enhances robustness against adversarial attacks of similar scale during training but also improves defense against stronger attacks.

\bibliographystyle{IEEEbib}
\bibliography{strings}

\begin{thebibliography}{10}

\bibitem{shafahi2019adversarial}
Ali Shafahi, Mahyar Najibi, Mohammad~Amin Ghiasi, Zheng Xu, John Dickerson,
  Christoph Studer, Larry~S Davis, Gavin Taylor, and Tom Goldstein,
\newblock ``Adversarial training for free!,''
\newblock {\em Advances in Neural Information Processing Systems}, vol. 32,
  2019.

\bibitem{cui2021learnable}
Jiequan Cui, Shu Liu, Liwei Wang, and Jiaya Jia,
\newblock ``Learnable boundary guided adversarial training,''
\newblock in {\em Proceedings of the IEEE/CVF International Conference on
  computer vision}, 2021, pp. 15721--15730.

\bibitem{zhang2018deep}
Ying Zhang, Tao Xiang, Timothy~M Hospedales, and Huchuan Lu,
\newblock ``Deep mutual learning,''
\newblock in {\em Proceedings of the IEEE Conference on Computer Cision and
  Pattern Recognition}, 2018, pp. 4320--4328.

\bibitem{huang2023boosting}
Bo~Huang, Mingyang Chen, Yi~Wang, Junda Lu, Minhao Cheng, and Wei Wang,
\newblock ``Boosting accuracy and robustness of student models via adaptive
  adversarial distillation,''
\newblock in {\em Proceedings of the IEEE/CVF Conference on Computer Vision and
  Pattern Recognition}, 2023, pp. 24668--24677.

\bibitem{cui2024decoupled}
Jiequan Cui, Zhuotao Tian, Zhisheng Zhong, Xiaojuan Qi, Bei Yu, and Hanwang
  Zhang,
\newblock ``Decoupled kullback-leibler divergence loss,''
\newblock {\em Advances in Neural Information Processing Systems}, vol. 37, pp.
  74461--74486, 2024.

\bibitem{jia2024improving}
Xiaojun Jia, Yong Zhang, Xingxing Wei, Baoyuan Wu, Ke~Ma, Jue Wang, and
  Xiaochun Cao,
\newblock ``Improving fast adversarial training with prior-guided knowledge,''
\newblock {\em IEEE Transactions on Pattern Analysis and Machine Intelligence},
  2024.

\bibitem{goodfellow2014explaining}
Ian~J Goodfellow, Jonathon Shlens, and Christian Szegedy,
\newblock ``Explaining and harnessing adversarial examples,''
\newblock {\em arXiv preprint arXiv:1412.6572}, 2014.

\bibitem{madry2017towards}
Aleksander Madry, Aleksandar Makelov, Ludwig Schmidt, Dimitris Tsipras, and
  Adrian Vladu,
\newblock ``Towards deep learning models resistant to adversarial attacks,''
\newblock {\em arXiv preprint arXiv:1706.06083}, 2017.

\bibitem{carlini2017towards}
Nicholas Carlini and David Wagner,
\newblock ``Towards evaluating the robustness of neural networks,''
\newblock in {\em 2017 IEEE Symposium on Security and Privacy (sp)}. Ieee,
  2017, pp. 39--57.

\bibitem{andriushchenko2020square}
Maksym Andriushchenko, Francesco Croce, Nicolas Flammarion, and Matthias Hein,
\newblock ``Square attack: a query-efficient black-box adversarial attack via
  random search,''
\newblock in {\em European Conference on Computer Vision}. Springer, 2020, pp.
  484--501.

\bibitem{croce2020minimally}
Francesco Croce and Matthias Hein,
\newblock ``Minimally distorted adversarial examples with a fast adaptive
  boundary attack,''
\newblock in {\em International Conference on Machine Learning}. PMLR, 2020,
  pp. 2196--2205.

\bibitem{rice2020overfitting}
Leslie Rice, Eric Wong, and Zico Kolter,
\newblock ``Overfitting in adversarially robust deep learning,''
\newblock in {\em International Conference on Machine Learning}. PMLR, 2020,
  pp. 8093--8104.

\bibitem{andriushchenko2020understanding}
Maksym Andriushchenko and Nicolas Flammarion,
\newblock ``Understanding and improving fast adversarial training,''
\newblock {\em Advances in Neural Information Processing Systems}, vol. 33, pp.
  16048--16059, 2020.

\bibitem{kim2021understanding}
Hoki Kim, Woojin Lee, and Jaewook Lee,
\newblock ``Understanding catastrophic overfitting in single-step adversarial
  training,''
\newblock in {\em Proceedings of the AAAI Conference on Artificial
  Intelligence}, 2021, vol.~35, pp. 8119--8127.

\bibitem{sriramanan2021towards}
Gaurang Sriramanan, Sravanti Addepalli, Arya Baburaj, et~al.,
\newblock ``Towards efficient and effective adversarial training,''
\newblock {\em Advances in Neural Information Processing Systems}, vol. 34, pp.
  11821--11833, 2021.

\bibitem{wong2020fast}
Eric Wong, Leslie Rice, and J~Zico Kolter,
\newblock ``Fast is better than free: Revisiting adversarial training,''
\newblock {\em arXiv preprint arXiv:2001.03994}, 2020.

\bibitem{jia2022prior}
Xiaojun Jia, Yong Zhang, Xingxing Wei, Baoyuan Wu, Ke~Ma, Jue Wang, and
  Xiaochun Cao,
\newblock ``Prior-guided adversarial initialization for fast adversarial
  training,''
\newblock in {\em European Conference on Computer Vision}. Springer, 2022, pp.
  567--584.

\bibitem{sriramanan2020guided}
Gaurang Sriramanan, Sravanti Addepalli, Arya Baburaj, et~al.,
\newblock ``Guided adversarial attack for evaluating and enhancing adversarial
  defenses,''
\newblock {\em Advances in Neural Information Processing Systems}, vol. 33, pp.
  20297--20308, 2020.

\bibitem{park2024dynamic}
Hyejin Park and Dongbo Min,
\newblock ``Dynamic guidance adversarial distillation with enhanced teacher
  knowledge,''
\newblock in {\em European Conference on Computer Vision}. Springer, 2024, pp.
  204--219.

\bibitem{krizhevsky2009learning}
Alex Krizhevsky, Geoffrey Hinton, et~al.,
\newblock ``Learning multiple layers of features from tiny images,''
\newblock 2009.

\bibitem{deng2009imagenet}
Jia Deng, Wei Dong, Richard Socher, Li-Jia Li, Kai Li, and Li~Fei-Fei,
\newblock ``Imagenet: A large-scale hierarchical image database,''
\newblock in {\em 2009 IEEE Conference on Computer Vision and Pattern
  Recognition}. Ieee, 2009, pp. 248--255.

\bibitem{jia2022boosting}
Xiaojun Jia, Yong Zhang, Baoyuan Wu, Jue Wang, and Xiaochun Cao,
\newblock ``Boosting fast adversarial training with learnable adversarial
  initialization,''
\newblock {\em IEEE Transactions on Image Processing}, vol. 31, pp. 4417--4430,
  2022.

\end{thebibliography}

\end{document}